\documentclass[letterpaper,10pt,conference]{ieeeconf}
\IEEEoverridecommandlockouts
\usepackage{iftex}
\ifXeTeX
  \usepackage{fontspec}
\else
\usepackage{tgtermes}
\fi

\usepackage{amsmath,amssymb}
\usepackage{dsfont}
\usepackage{booktabs}
\usepackage{cite}
\usepackage{graphicx}
\usepackage{xurl}
\usepackage{xspace}
\usepackage{xcolor}
\usepackage{multirow}

\newcommand{\method}{TacSushi\xspace}

\title{\method{}: Tactile-Grounded World--Action Modeling \\
for Dexterous Sushi Manipulation}
\newif\ificraanonymous

\ificraanonymous
  \author{Anonymous Authors}
\else
  \author{%
  \textbf{Haodi Hu}$^{1, 2}$$^\dagger$, 
  \textbf{Kaen Kogashi}$^{2, 3}$$^\dagger$,
  \textbf{Toshiaki Koike-Akino}$^2$ \\
  $^\dagger$ equal contribution \\
  $^1$University of Southern California, USA \\
  $^2$Mitsubishi Electric Research Laboratories (MERL), USA \\
  $^3$Mitsubishi Electric, Japan
  }
\fi

\begin{document}
\bstctlcite{tacsushi:compactauthors}
\maketitle
\thispagestyle{empty}
\pagestyle{empty}

\begin{abstract}
Dexterous food manipulation requires control under deformation, occlusion, and uncertain contact. We present \method{}, a tactile-grounded, Cosmos3-based world--action policy that learns from recorded future consequences while acting on current observations. The backbone encodes current RGB, language, and hand state, and feature-wise gated fusion incorporates fingertip tactile features into the action representation. During training, a decoder conditioned on demonstrated action chunks predicts logged future visual observations, task progress, relative contact risk, and tactile summaries; this decoder is removed at deployment. Failed trials provide consequence supervision, but their actions are excluded from imitation. We train \method{} on 340 successful and 50 failed real-robot trials and compare six methods in 600 separate rollouts across three in-distribution tasks and two out-of-distribution ingredient variants. To assess food quality beyond a single geometric threshold, we score terminal outcomes using an anchored visual-quality protocol that equally weights five human ratings and three vision-language-model ratings per rollout. Full \method{} achieves 68.3\% average in-distribution success and 37.5\% out-of-distribution success, compared with 36.7\%/10.0\% without future-consequence supervision and 25.0\%/17.5\% with direct tactile concatenation in place of gated fusion. These comparisons support complementary benefits of feature-wise gated tactile fusion and training-only predictive supervision.
\end{abstract}

\section{Introduction}
Food manipulation exposes a central limitation of vision-dominant robot policies: an object can appear acceptable while local contact is already becoming unstable. Sticky rice changes shape and adheres to the fingers, seaweed bends and slips, and fragile toppings can be displaced or damaged by small errors in force or timing. Ingredient properties also vary between samples~\cite{wang2021foodhandling}, and haptic cues help guide food acquisition~\cite{bhattacharjee2019feeding}. Sushi preparation is therefore a useful test of whether a policy can relate current contact to the likely consequences of its actions, especially when the hand occludes the contact region.

Research on deformable manipulation has developed simulation benchmarks, differentiable physics, and learned dynamics for cloth and elasto-plastic materials~\cite{lin2021softgym,huang2021plasticinelab,shi2024robocraft}. Learning-based granular manipulation also exploits material motion for control~\cite{hu2025grain,hu2025diffusivegrain}. These studies motivate learning the consequences of robot--material interaction, although their indirect manipulation through granular flow differs from direct fingertip manipulation of food. Food-handling systems have combined vision and touch for specialized acquisition or cutting operations~\cite{bhattacharjee2019feeding,SashimiBot}. Dexterous preparation of real food remains difficult: the controller must coordinate several fingertips, tolerate adhesion and irreversible deformation, and transfer across ingredients that differ in both appearance and mechanics. This setting motivates learning contact-sensitive behavior from real demonstrations.

At the policy level, generalist vision--language--action models provide reusable semantic and visual priors~\cite{brohan2022rt1,oneill2023openx,brohan2023rt2,octo2024,kim2024openvla,nvidia2026gr00t}, and action-chunking approaches improve temporal consistency~\cite{zhao2023act,chi2023diffusionpolicy}. A complementary line of work learns how actions transform observations. Latent world models support planning or policy learning through imagined trajectories~\cite{hafner2019planet,hafner2020dreamer}, while visual-foresight methods use predicted frames in an online control loop~\cite{finn2017deepvisualforesight,ebert2018visualforesight}. For deformable food, online model rollouts can make control depend on predictions where long-horizon geometry and contact are difficult to model. This motivates using recorded future consequences to shape the policy representation during training, without requiring predicted futures during execution.

The distinction between deployed observations and privileged training signals is especially important for touch. Tactile sensing supplies local evidence that external vision may miss~\cite{li2020tactileinformation,lambeta2020digit} and has supported regrasp selection, grasp-outcome prediction, and state estimation under occlusion~\cite{calandra2018more,sodhi2021tactile}. Cross-modal objectives learn shared visual--tactile representations~\cite{lee2019making,li2019touchvision}, while predictive tactile models have enabled model-predictive control and action-conditioned slip forecasting~\cite{tian2019manipulationfeel,mandil2022action}. Recent imitation policies use current vision and touch, as in TACT~\cite{murooka2025tact}, or add future-tactile prediction, as in ViTacFormer~\cite{heng2025vitacformer}. Together, these results suggest two complementary roles for tactile information, but do not fully resolve how current touch, which can guide the deployed policy, should interact with future visual and contact outcomes that exist only in the training record.

\method{} addresses this question by separating training-time consequence prediction from deployment-time control in a Cosmos3-based policy~\cite{nvidia2026cosmos3}. A feature-wise gate injects current fingertip features into the visual--language--proprioceptive latent that drives the action head. The same fused latent also supports a training-only consequence decoder. Current touch can therefore modify the command directly, while future observations supervise features that help predict task evolution. The deployed controller does not consume predicted outcomes. Our contribution lies in coupling this deployed gated-fusion path with training-only, multi-target consequence supervision, rather than in tactile fusion, future prediction, or action conditioning individually.

The consequence decoder receives the demonstrated action chunk aligned with each training window. Failed trajectories retain their logged visual, contact, and outcome targets, while an action-loss mask excludes failed commands from imitation. These logged action--consequence pairs supervise the shared representation; the decoder is not used as a causal simulator or online planner.

Our experiments address two questions: whether feature-wise gated tactile fusion improves control over direct tactile concatenation, and whether future-consequence supervision improves the deployed policy even though the auxiliary decoder is removed at deployment. We vary both factors in a matched four-variant study and examine transfer to held-out ingredients. To assess heterogeneous food outcomes, we use a human--VLM protocol that assigns each rollout a success label from its terminal appearance. The principal contributions are:
\begin{enumerate}
    \item a Cosmos3-based world--action policy that gives current fingertip touch a direct, feature-wise gated path to action generation while reserving future consequences for training-only representation learning;
    \item a multi-target objective conditioned on demonstrated actions that learns from future visual, tactile, progress, and contact-risk outcomes, with failure-aware masking that preserves consequence supervision while excluding failed actions from imitation; and
    \item a real-robot study comparing tactile-fusion strategies and privileged consequence supervision, with evaluation on held-out ingredients and a multi-rater human--VLM protocol for assessing food outcomes.
\end{enumerate}

\section{Method}

\begin{figure}[t]
    \centering
    \includegraphics[width=\linewidth]{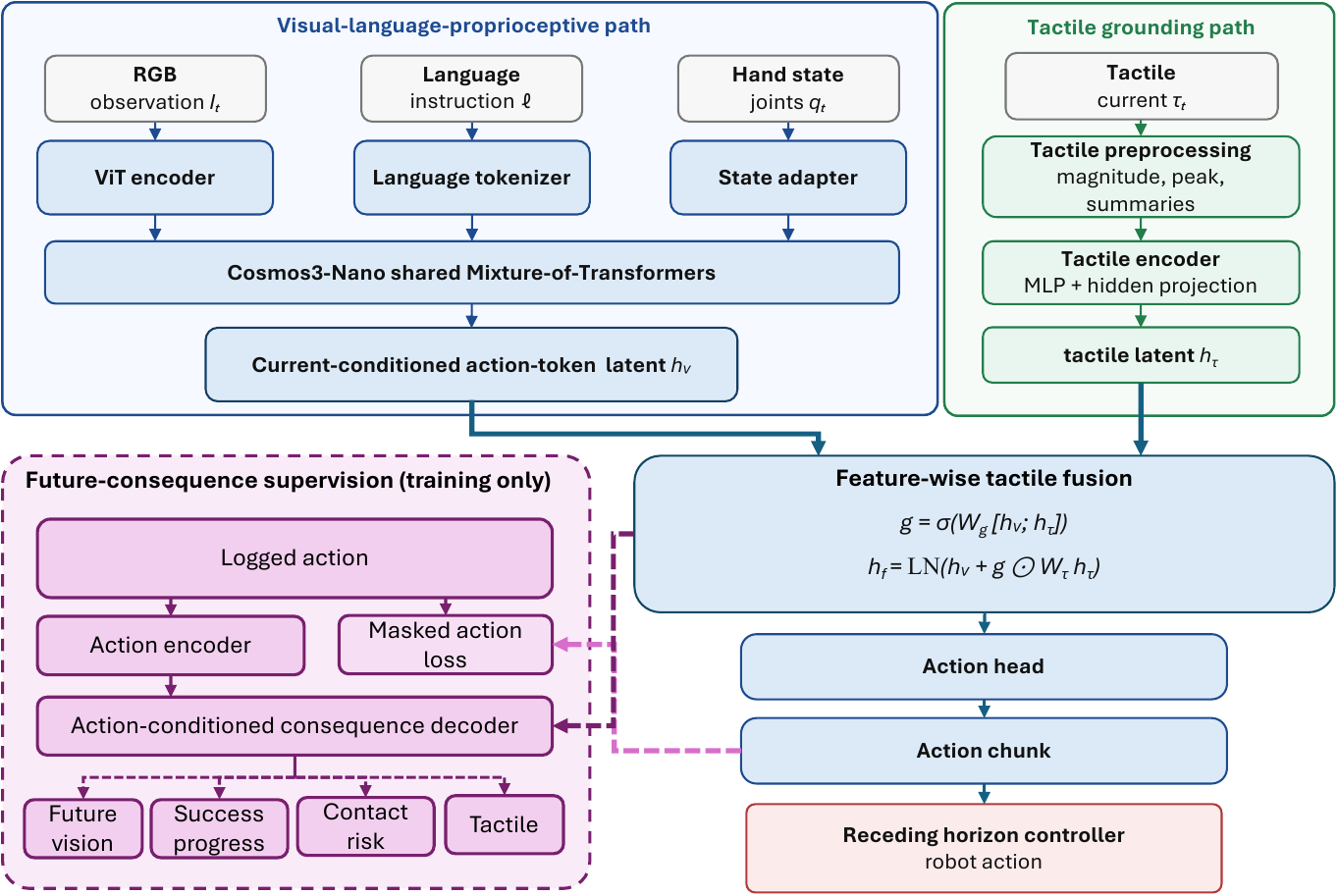}
    \caption{Detailed architecture and train--deploy asymmetry. Current visual--language--proprioceptive and tactile latents are fused into $h_f$, which independently feeds the deployed action head and a training-only consequence decoder. The decoder combines $h_f$ with the demonstrated-action embedding to predict future vision, success progress, contact risk, and a tactile summary (abbreviated ``Tactile'' in the box). The masked action loss excludes failed commands from imitation, while their recorded consequences remain supervised. Solid blue denotes the path used in training and deployment; dashed purple denotes privileged training supervision. Only the action path reaches the receding-horizon execution controller.}
    \label{fig:network-detail}
\end{figure}

\method{} learns from future consequences during training and acts on current sensing at deployment. The policy observes current RGB, language, hand state, and fingertip touch and predicts an actuator-target chunk for guarded receding-horizon execution. During training, an action-conditioned decoder regularizes the fused policy latent by predicting logged visual, tactile, and task outcomes. We use \emph{world--action modeling} in this representation-learning sense: the model predicts observable consequences of the demonstrated action chunk given the current observation. At deployment, the controller in Fig.~\ref{fig:network-detail} executes a chunk prefix and re-observes without querying the consequence decoder.

\subsection{System, Teleoperation, and Data}
\method{} controls a four-finger Shadow Dexterous Hand Lite with a Shadow Tactile Fingertip (STF) sensor on each fingertip, including the thumb. Each STF contains 17 three-axis taxels, giving 68 taxels and 204 raw scalar tactile channels across the hand. The hand has 13 actuators and 16 observed joints, including three passive joints. Each action therefore specifies 13 actuator-position targets; the passive joints are observed but are not independently commanded. The policy predicts chunks of $H_a=16$ action steps, where $H_a$ denotes the temporal prediction horizon, not the action dimension. A Cosmos3-Nano backbone encodes RGB, language, and the 16-joint hand state; a compact tactile encoder and learned gate incorporate current contact into the action latent. Tactile preprocessing reduces each taxel's three-axis measurement to its Euclidean magnitude, yielding 68 physical magnitude features, and computes peaks and causal summaries from samples available at or before $t$, as shown in Fig.~\ref{fig:network-detail}.

\begin{figure}[t]
    \centering
    \includegraphics[width=\linewidth]{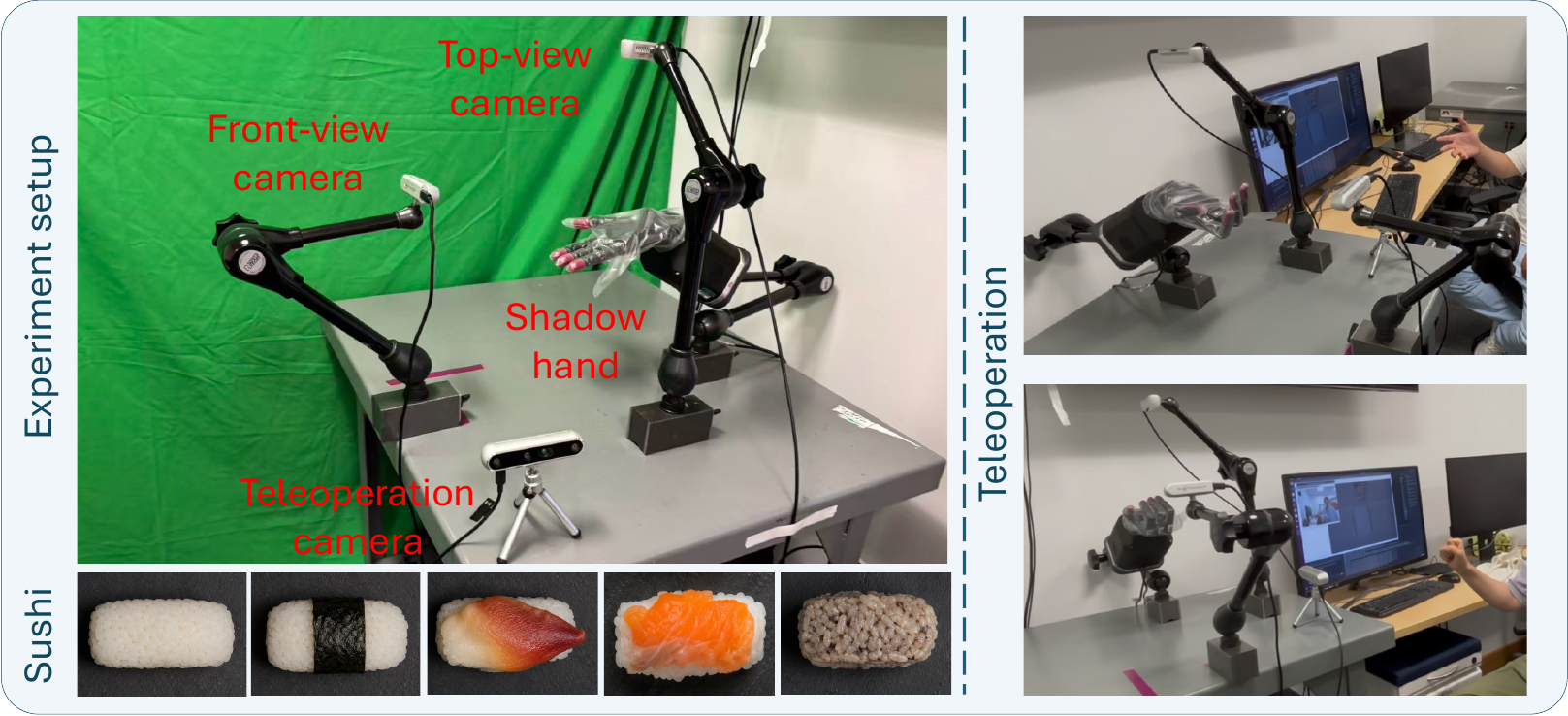}
    \caption{Platform and bare-hand collection. MediaPipe landmarks drive the tactile-equipped Shadow Hand; workspace RGB, robot state/targets, and touch are logged synchronously.}
    \label{fig:exp_setup}
\end{figure}

For data collection, a RealSense tracking camera supplies 30\,Hz operator-view RGB to MediaPipe Hand Landmarker~\cite{zhang2020mediapipehands}. Estimated human-hand joints are retargeted to 13 robot actuator commands using fixed gains and offsets; the three passive robot joints do not receive independent targets. At up to 10\,Hz, targets are clipped, low-pass filtered ($\alpha=0.30$), step-limited to 0.10\,rad, smoothly interpolated, and checked for self-collision using the hand's joint configuration. This bare-hand interface lets an operator demonstrate contact-rich motions without an exoskeleton or instrumented glove. Figure~\ref{fig:exp_setup} shows the interface and collection setup.

Separate front-view and top-view cameras record the workspace while the 16-joint robot state, 13-D actuator targets, and fingertip tactile measurements are logged synchronously at 30\,Hz. The operator-view stream supports teleoperation but is excluded from policy observations.

Table~\ref{tab:dataset_v2} summarizes the corpus. Successful trials supply action and positive-progress targets. Failed teleoperation supplies recorded adverse visual and contact consequences and a zero success-outcome target, but its action-imitation mask is zero. 

\begin{table}[t]
\centering
\caption{Training-corpus statistics; duration and frames per workspace-camera stream are averaged over successful trials.}
\label{tab:dataset_v2}
\resizebox{\columnwidth}{!}{%
\begin{tabular}{lcccc}
\toprule
Statistic & Rice ball & Seaweed & Surf clam & Overall \\
\midrule
Successful trials & 114 & 113 & 113 & 340 \\
Failed trials & 16 & 17 & 17 & 50 \\
Avg. duration (s) & 74 & 117 & 71 & 87.3 \\
Avg. RGB frames & 2,220 & 3,510 & 2,130 & 2,619 \\
\bottomrule
\end{tabular}}
\end{table}

\subsection{Tactile-Gated Action and Training-Only Consequence Prediction}
\textbf{Deployed action path.}
At control step $t$, the policy receives only the current observation $x_t=(I_t,\ell,q_t,\tau_t)$, comprising RGB $I_t$, language instruction $\ell$, the observed joint configuration $q_t\in\mathbb R^{16}$ (including the three passive joints), and fingertip measurements $\tau_t$. The Cosmos3-Nano shared Mixture-of-Transformers~\cite{nvidia2026cosmos3} encodes the first three inputs and returns the current-conditioned action-token representation $h_v=f_\theta(I_t,\ell,q_t)$; $h_v$ is distinct from the future-vision decoder output. In parallel, a compact tactile encoder uses measurements available at $t$ to produce $h_\tau=E_\psi(\tau_t)$.

The two representations are combined by the feature-wise gate in Fig.~\ref{fig:network-detail}:
\begin{equation}
\begin{aligned}
g_t&=\sigma\!\left(W_g[h_v;h_\tau]\right),\\
h_f&=\operatorname{LN}\!\left(h_v+g_t\odot W_\tau h_\tau\right).
\end{aligned}
\label{eq:gated_fusion}
\end{equation}
Here, $W_\tau h_\tau$ projects touch into the backbone feature space, and $g_t\in[0,1]^d$ scales each feature of the tactile residual, where $d$ is the hidden dimension. The action head predicts $\hat{\mathbf a}_t=A_\phi(h_f)\in\mathbb R^{H_a\times13}$, a 16-step chunk with one 13-D actuator-target vector per step. The controller executes a prefix before re-observing. This path supplies all deployed action predictions.

\textbf{Training-only consequence path.}
During training, the fused state $h_f$ also branches to a separate consequence decoder. An action encoder embeds the demonstrated actuator-target chunk $\mathbf a_t^\star\in\mathbb R^{H_a\times13}$ together with the current 16-joint observation as $e_{a,t}^\star=E_a(\mathbf a_t^\star,q_t)$. Conditioned on both the current fused state and the action that was actually demonstrated, the decoder produces
$(\hat I_t,\hat v_t,\hat r_t,\hat b_t)=U_\omega(h_f,e_{a,t}^\star)$, corresponding to future vision, success progress, relative contact risk, and a tactile summary. The future-vision output uses the Cosmos generation objective; the remaining readouts regress their respective targets.

Decoder losses back-propagate through $h_f$ and the selected backbone adapters, so the action and consequence objectives shape a shared tactile-grounded representation. The action head does not depend on consequence-decoder outputs: neither $e_{a,t}^\star$ nor a decoder prediction is passed to $A_\phi$. The action encoder and consequence decoder are removed at deployment.

\textbf{Aligned training windows.}
Each example is anchored at time $t$, with actions and future observations aligned over the same physical interval. It contains the current observation $x_t$, demonstrated action chunk $\mathbf a_t^\star$, future RGB sequence $I_t^\star=I_{t+1:t+H_I}$, progress target $y_t^{\rm value}$, contact-risk target $y_t^{\rm risk}$, and tactile/phase target $b_t^{\rm tac}$. Actuator targets are resampled to $H_a=16$ action steps and RGB to $H_I=33$ frames with aligned endpoints. The decoder therefore predicts the consequences of the action interval on which it is conditioned, despite the different sampling rates. Future observations and annotations serve only as training targets.

The four targets describe complementary consequences. Future RGB captures visible motion and deformation; the tactile summary records contact evolution and brief loading events; progress indicates task advancement; and contact risk marks adverse interaction. None is treated as a sensor or policy input at test time. Instead, their gradients encourage $h_f$ to encode outcome-relevant aspects of the current observation.

\textbf{Failure-aware objective.}
The demonstrated chunk conditions the decoder on the command associated with each logged consequence window. For failed windows, that chunk remains valid conditioning for the recorded adverse consequences, but its imitation mask is zero. Thus failed trajectories supervise the consequence objectives without treating failed commands as policy targets. The decoder is trained only on logged action--consequence pairs and is not used for counterfactual prediction. Formally,
\begin{equation}
\mathcal L_{\rm train}
=m_t\mathcal L_{\rm action}
+\lambda_{\rm vis}\mathcal L_{\rm future\text{-}vis}
+\mathcal L_{\rm aux},
\label{eq:wam_loss}
\end{equation}
where $m_t=1$ for successful demonstrations and $0$ for failed demonstrations. Only the imitation term is masked. $\mathcal L_{\rm aux}$ contains the progress, risk, tactile-summary, and risk--summary consistency losses. We train the task-specific action, tactile, fusion, and consequence modules together with selected backbone adapters, while keeping most pretrained parameters frozen.

\subsection{Privileged Tactile and Outcome Targets}
For a window beginning at sample $t$, let
$\boldsymbol\tau_s^{\rm mag}\in\mathbb R^{K}$ denote the tactile-magnitude vector at sample $s$, where $K=68$ is the number of per-taxel magnitude channels used for training (17 per fingertip across four STF sensors). Each channel is the Euclidean magnitude of one taxel's three-axis measurement. For a consequence horizon of $H_c$ tactile samples, define $\mathcal S_t=\{t+1,\ldots,t+H_c\}$ and summarize the strongest instantaneous contact by $u_s=\max_{k\in\{1,\ldots,K\}}\tau_{s,k}^{\rm mag}$. If the complete episode contains $T$ samples, its normalized window endpoint is $\rho_t=\min\{(t+H_c)/T,1\}$. The episode-level quantile $Q_{\alpha}(u_{1:T})$ supplies a relative contact scale, where $\alpha\in(0,1)$ is the selected quantile level.

With a small $\varepsilon>0$ preventing division by zero, the relative peak-contact target and risk-adjusted progress target are
\begin{align}
y_t^{\rm risk}
 &=\operatorname{clip}\!\left(
   \frac{\max_{s\in\mathcal S_t}u_s}
        {Q_{\alpha}(u_{1:T})+\varepsilon},0,1\right), \nonumber\\
y_t^{\rm value}
 &=m_t\rho_t(1-y_t^{\rm risk}).
\label{eq:aux_targets}
\end{align}
Here, $\operatorname{clip}(\cdot,0,1)$ bounds the peak ratio, and $m_t=1$ for a successful demonstration and $0$ for a failed one, matching the imitation mask in Eq.~\eqref{eq:wam_loss}. In our experiments, $H_c=33$ samples (1.10\,s at 30\,Hz), $\alpha=0.90$, and $\varepsilon=10^{-6}$; the tactile and RGB consequence windows share the same endpoint. Thus $y_t^{\rm risk}$ measures peak contact relative to the episode's contact distribution, while $y_t^{\rm value}$ weights normalized task phase by successful outcome and relative contact risk.

The tactile target retains more information than either scalar:
\begin{equation}
\begin{aligned}
\bar{\boldsymbol\tau}_t^{\rm mag}
  &=\frac{1}{H_c}\sum_{s\in\mathcal S_t}
    \boldsymbol\tau_s^{\rm mag},\\
b_t^{\rm tac}
  &=\left[
    \boldsymbol\tau_{t+H_c}^{\rm mag};
    \bar{\boldsymbol\tau}_t^{\rm mag};
    y_t^{\rm risk};
    \rho_t
    \right]\in\mathbb R^{2K+2}.
\end{aligned}
\label{eq:tactile_summary}
\end{equation}
Its first $K$ entries describe contact at the window endpoint, the next $K$ describe mean contact over the interval, and the final two record relative peak contact and normalized episode phase. With $K=68$, this gives a 138-D target. Endpoint and mean contact distinguish residual, sustained, and transient loading that a single peak cannot represent.

With $[\hat b_t]_{\rm risk}$ denoting the predicted risk coordinate of the summary, the nonvisual auxiliary loss is
\begin{equation}
\begin{aligned}
\mathcal L_{\rm aux}={}&
\lambda_{\rm value}(\hat v_t-y_t^{\rm value})^2
+\lambda_{\rm risk}(\hat r_t-y_t^{\rm risk})^2\\
&+\frac{\lambda_{\rm tac}}{2K+2}
  \|\hat b_t-b_t^{\rm tac}\|_2^2
+\lambda_{\rm cons}
  (\hat r_t-[\hat b_t]_{\rm risk})^2.
\end{aligned}
\label{eq:aux_loss}
\end{equation}
The four $\lambda$ coefficients are fixed loss weights shared across the controlled variants. The consistency term aligns the scalar risk prediction with the risk coordinate of the tactile summary. Dividing the summary loss by $2K+2$ normalizes it by the target dimension. Future samples, the episode quantile, and the episode outcome are used only to construct training targets and are unavailable to the deployed action head.

\begin{figure}[t]
    \centering
    \includegraphics[width=\linewidth]{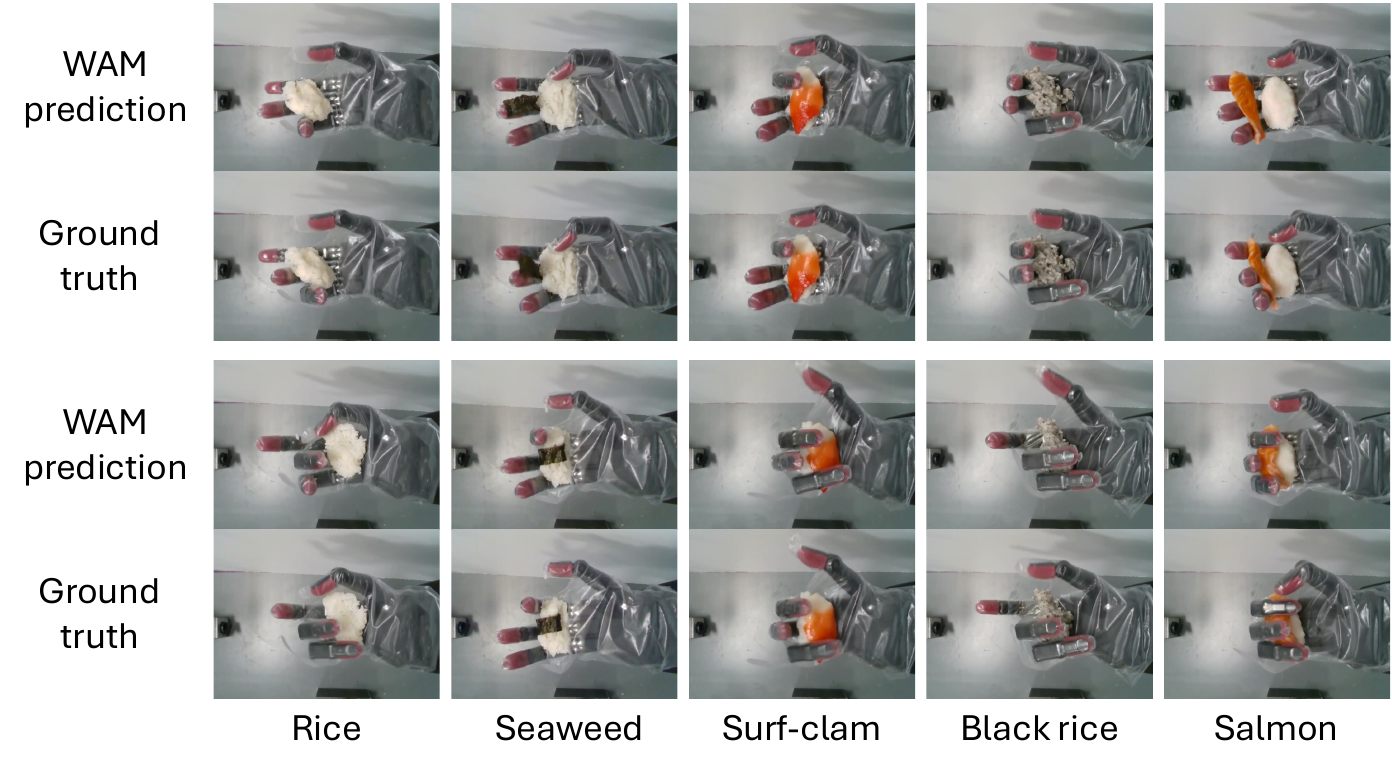}
    \caption{Full \method{} future-vision predictions and synchronized ground truth for all five food categories. Each row pair shows predictions above ground truth. Table~\ref{tab:wam-eval} reports fixed-set reconstruction errors; this readout is not used by deployed control.}
    \label{fig:wam-prediction}
\end{figure}

\subsection{Training, Deployment, and Evaluated Methods}
\label{sec:training-variants}
Training jointly optimizes all enabled terms in Eq.~\eqref{eq:wam_loss}. Successful windows supervise action generation, and every valid window supervises the enabled consequence objectives. The failure mask $m_t$ affects only imitation. Task-specific modules are trained together with selected Cosmos3 adapters; the remaining pretrained parameters are frozen. At deployment, current RGB, the 16-joint state, touch, and language enter the solid-blue path in Fig.~\ref{fig:network-detail}. The server returns one $H_a\times13$ action chunk. The receding-horizon controller clamps the actuator targets, checks the associated hand configuration for self-collision, executes the first $K_{\rm exec}\leq H_a$ steps, and re-observes before requesting the next chunk, where $K_{\rm exec}$ is the number of steps committed per update. We used $K_{\rm exec}$ = 12 in this work.

\textbf{Controlled Cosmos3 variants (M3--M6).}
We vary two factors: the tactile-fusion strategy and the use of training-only consequence supervision. All four variants receive current tactile observations. M3 uses direct tactile concatenation without consequence supervision (gate$-$, auxiliary$-$); M4 uses the same concatenation with consequence supervision (gate$-$, auxiliary$+$); M5 uses feature-wise gated tactile fusion without consequence supervision (gate$+$, auxiliary$-$); and M6 is full \method{}, with gated fusion and consequence supervision (gate$+$, auxiliary$+$).

The gate$-$ variants retain the tactile encoder and directly concatenate its output with the visual--language--proprioceptive representation. The gate$+$ variants use the gated residual in Eq.~\eqref{eq:gated_fusion}. Removing consequence supervision sets the future-vision and all other consequence-loss weights to zero while preserving the deployed tactile pathway. M3--M6 share the training data, data order, number of optimization steps, optimizer schedule, action-loss settings, and guarded executor. M3/M4 and M5/M6 compare consequence supervision at a fixed fusion architecture; M3/M5 and M4/M6 compare gated fusion with direct concatenation at a fixed supervision setting. All four retain the pretrained visual generator for offline evaluation, but only M4 and M6 adapt it using future-consequence targets. None uses the generator during deployed control.

\textbf{System-level baselines (M1--M2).}
M1 is GR00T without tactile input. M2 augments GR00T with current touch: an MLP embeds the tactile portion of the state vector, and a sigmoid gate injects that embedding into the state representation before action prediction. M1--M2 provide cross-backbone system comparisons.

\begin{figure*}[tbp]
    \centering
    \includegraphics[width=0.99\textwidth]{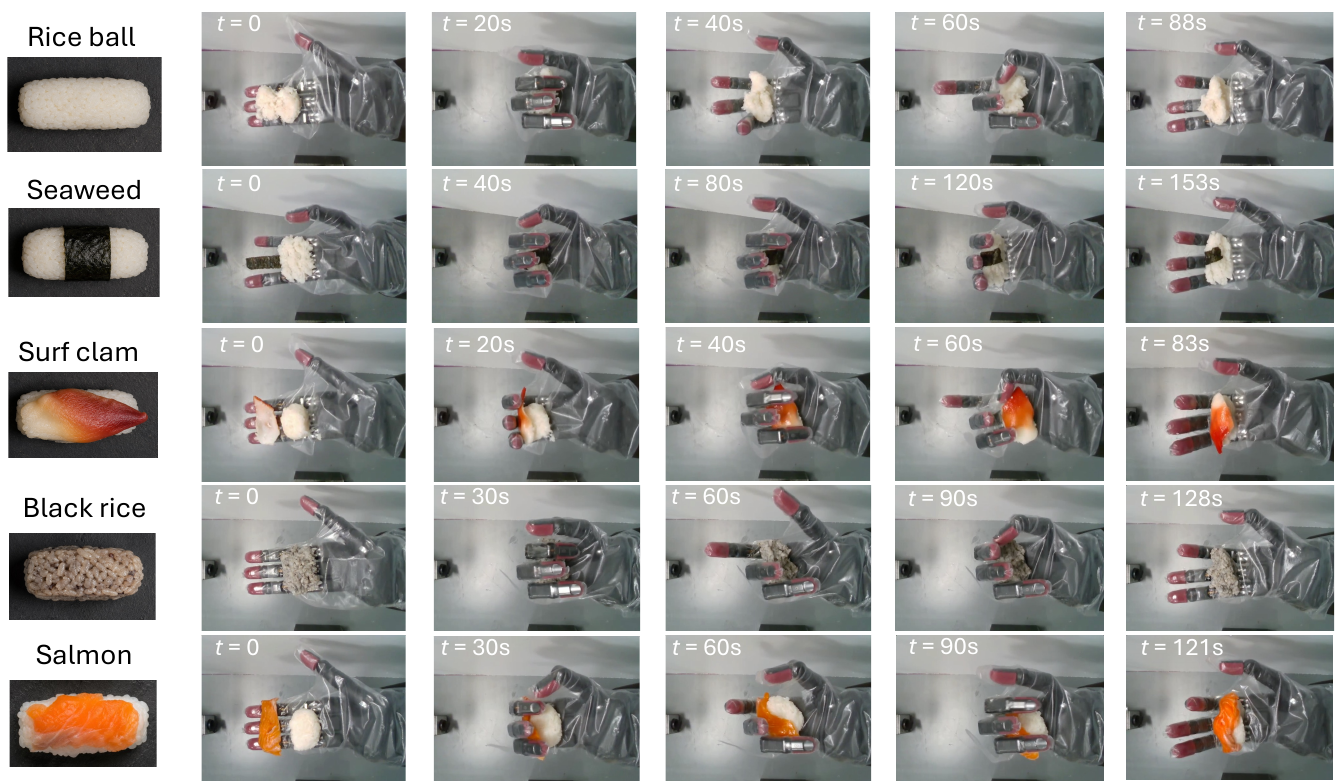}
    \caption{Representative successful executions on all five conditions. The first three rows show in-distribution rice-ball shaping, seaweed wrapping, and surf-clam placement; the final two show held-out black-rice and salmon ingredient variants. Each row begins with the desired appearance and then traces top-camera frames from the initial state to the terminal outcome.}
    \label{fig:id-executions}
\end{figure*}

\vspace{-0pt}

\section{Experiments}
\subsection{Tasks and Rollout Protocol}
We evaluate the six methods defined in Sec.~\ref{sec:training-variants} on three in-distribution (ID) tasks---rice-ball shaping, seaweed wrapping, and surf-clam placement---and two held-out out-of-distribution (OOD) ingredient variants, black rice and salmon. Each method--task condition includes 20 trials, giving 600 real-robot rollouts separate from the training corpus. Every method uses the platform and guarded execution interface in Fig.~\ref{fig:exp_setup}.

Rice-ball shaping compacts loose rice into a self-supporting oval portion; seaweed wrapping closes the sheet around the rice; and surf-clam placement centers and retains the topping without visible crushing. Black rice preserves the shaping objective while changing the ingredient, whereas salmon replaces the seen surf-clam topping while preserving the placement objective. Neither held-out ingredient is used for task-specific adaptation. 
Figure~\ref{fig:id-executions} shows intermediate behavior omitted by a terminal score; its final frames illustrate the visual outcomes judged by the protocol below.

\begin{table*}[tbp]
\centering
\caption{Human and AI-ensemble visual-quality scores on the five-point scale (mean $\pm$ standard error of the mean, SEM; 20 rollouts per method--category cell). Category MAE averages the absolute human--AI differences between the six method means; lower is better. M1--M6 are defined in Sec.~\ref{sec:training-variants}.}
\label{tab:user-survey}
\resizebox{\textwidth}{!}{%
\setlength{\tabcolsep}{6pt}
\begin{tabular}{ll cccccc c}
\toprule
\textbf{Category} & \textbf{Group} & \textbf{M1} & \textbf{M2} & \textbf{M3} & \textbf{M4} & \textbf{M5} & \textbf{M6} & \textbf{MAE} ($\downarrow$) \\
\midrule
\multirow{2}{*}{Rice ball} 
 & Human & $1.96 \pm 0.29$ & $2.66 \pm 0.32$ & $1.80 \pm 0.26$ & $2.22 \pm 0.31$ & $2.02 \pm 0.29$ & $3.34 \pm 0.28$ & \multirow{2}{*}{0.07} \\
 & AI & $2.02 \pm 0.26$ & $2.65 \pm 0.35$ & $1.91 \pm 0.28$ & $2.28 \pm 0.27$ & $2.18 \pm 0.31$ & $3.31 \pm 0.26$ & \\
\cmidrule(lr){1-9}
\multirow{2}{*}{Seaweed} 
 & Human & $1.42 \pm 0.19$ & $2.26 \pm 0.31$ & $1.68 \pm 0.23$ & $1.68 \pm 0.23$ & $2.22 \pm 0.31$ & $2.92 \pm 0.31$ & \multirow{2}{*}{0.09} \\
 & AI & $1.62 \pm 0.18$ & $2.42 \pm 0.29$ & $1.77 \pm 0.22$ & $1.76 \pm 0.22$ & $2.28 \pm 0.29$ & $2.93 \pm 0.34$ & \\
\cmidrule(lr){1-9}
\multirow{2}{*}{Surf clam} 
 & Human & $1.92 \pm 0.28$ & $2.78 \pm 0.32$ & $1.46 \pm 0.19$ & $1.88 \pm 0.28$ & $2.50 \pm 0.28$ & $3.20 \pm 0.29$ & \multirow{2}{*}{0.10} \\
 & AI & $2.04 \pm 0.24$ & $2.76 \pm 0.30$ & $1.62 \pm 0.18$ & $2.06 \pm 0.26$ & $2.58 \pm 0.30$ & $3.24 \pm 0.28$ & \\
\cmidrule(lr){1-9}
\multirow{2}{*}{Salmon (OOD)} 
 & Human & $1.44 \pm 0.17$ & $1.58 \pm 0.23$ & $1.38 \pm 0.14$ & $1.98 \pm 0.24$ & $1.52 \pm 0.19$ & $2.26 \pm 0.34$ & \multirow{2}{*}{0.13} \\
 & AI & $1.62 \pm 0.18$ & $1.77 \pm 0.22$ & $1.47 \pm 0.13$ & $2.10 \pm 0.26$ & $1.58 \pm 0.18$ & $2.42 \pm 0.30$ & \\
\cmidrule(lr){1-9}
\multirow{2}{*}{Black rice (OOD)} 
 & Human & $1.30 \pm 0.16$ & $1.58 \pm 0.23$ & $1.36 \pm 0.14$ & $1.52 \pm 0.22$ & $1.54 \pm 0.19$ & $2.27 \pm 0.29$ & \multirow{2}{*}{0.11} \\
 & AI & $1.52 \pm 0.13$ & $1.78 \pm 0.22$ & $1.47 \pm 0.13$ & $1.50 \pm 0.18$ & $1.56 \pm 0.18$ & $2.18 \pm 0.29$ & \\
\bottomrule
\end{tabular}%
}
\end{table*}

\subsection{Human--VLM Outcome Evaluation and User Survey}
Acceptable food outcomes vary with ingredient and task, making a single geometric or force threshold unsuitable for judging shape, wrapping, topping placement, and visible damage. A fixed geometric tolerance can reject natural variation yet overlook obvious failure. We therefore combine a photo-based human survey with VLM evaluation under a shared, anchored five-point visual-quality scale. Task-specific criteria include compact, self-supporting rice; a wrap that remains closed; a centered and retained topping; and no visible crushing.

\textbf{Survey protocol.}
Fifty voluntary, uncompensated participants provided written consent and evaluated terminal images from the 600 robot rollouts. Each image was entered five times into the assignment pool, producing 3,000 image-rating assignments and five human ratings per rollout. Each participant rated 60 images with 12 randomly assigned images from each of the five food categories. As a result, each category contributed 120 unique rollout images and 600 rating assignments. Before rating, participants read brief guidance on desirable sushi presentation: an oval rice portion, balanced topping placement, gentle pressing, and avoidance of over-compression. The anchored five-point scale was 5, excellent/visually perfect; 4, good/well-prepared; 3, acceptable/neutral; 2, poor/slightly messy; and 1, very poor/disorganized. The study underwent institutional operations review, with the institution withheld for double-blind review. Participation required approximately 15 minutes, and responses were anonymized. Participants were told that pictured sushi could have been produced by a chef, a conveyor-belt machine, or a dexterous robotic hand, but individual images did not display source labels.

\textbf{Trial-level composite score.}
GPT-5.6, Claude Opus 4.8, and Gemini 3.5 independently score the same terminal images using a common prompt and the survey rubric. The input to the 3 VLMs is a survey with a prompt that describes the general evaluation rule and instructions as the same as used for the human survey. Each rollout therefore receives eight equally weighted ratings: five from humans and one from each of the three VLMs. For trial $j$, let $\mathcal H_j$ be its set of five human raters, $r_{j,h}^{\rm human}$ the rating from human $h$, and $r_{j,k}^{\rm AI}$ the rating from VLM $k\in\{1,2,3\}$. The human mean $\bar r_j^{\rm human}$, AI-ensemble mean $\bar r_j^{\rm AI}$, and composite score $s_j$ are
\begin{equation}
\begin{aligned}
\bar r_j^{\rm human}
&=\frac{1}{|\mathcal H_j|}\sum_{h\in\mathcal H_j}r_{j,h}^{\rm human},\\
\bar r_j^{\rm AI}
&=\frac{1}{3}\sum_{k=1}^{3}r_{j,k}^{\rm AI},\\
s_j
&=\frac{5}{8}\bar r_j^{\rm human}
  +\frac{3}{8}\bar r_j^{\rm AI}.
\end{aligned}
\label{eq:visual_adjudication}
\end{equation}
Every individual rating contributes $1/8$ of the composite score. Consequently, the five-rater human mean carries a weight of $5/8$ (62.5\%), and the three-model AI mean carries $3/8$ (37.5\%). The weighting is equal across individual ratings, not across the two evaluator groups. Trial $j$ is successful if and only if $s_j\geq3$, with 3 marking acceptable task completion. Each success label is computed from the eight ratings of that specific rollout, not from a method-level or category-level mean.

\begin{figure*}[tbp]
    \centering
    \includegraphics[width=0.96\textwidth]{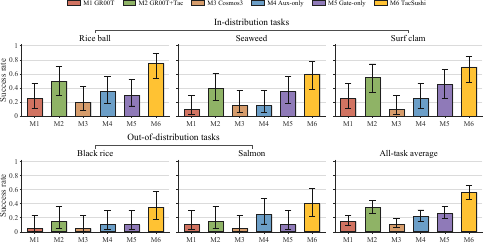}
    \caption{Real-robot success under the trial-level human--VLM visual-outcome rule. Error bars are two-sided 95\% Wilson intervals using $n=20$ per task and $n=100$ for the descriptive pooled panel; M1--M6 are defined in Sec.~\ref{sec:training-variants}.}
    \label{fig:tacsushi-results}
\end{figure*}

Success rates average the binary labels over 20 trials per method--task condition. Task-level Wilson intervals use the integer success count with $n=20$. The all-task panel reports a descriptive pooled proportion and Wilson interval for the prespecified five-task mixture ($n=100$), rather than a common success probability for all tasks. Intervals in Fig.~\ref{fig:tacsushi-results} reflect rollout sampling conditional on one training run and checkpoint per method and the fixed labels from Eq.~\eqref{eq:visual_adjudication}. They do not capture training-seed variability or uncertainty from the evaluators and human raters.

\textbf{Human--AI score comparison.}
Table~\ref{tab:user-survey} reports the human and AI-ensemble scores separately by method and food category. Each cell gives the mean and SEM of the corresponding rollout-level means in Eq.~\eqref{eq:visual_adjudication} over 20 trials. The final column averages the absolute differences between the six paired human and AI method means within each category, in rating points. It therefore describes agreement between aggregate scores rather than ratings of individual images.

Full \method{} (M6) receives the highest mean rating from both evaluator groups in every category. Its human scores are $3.34\pm0.28$ for rice ball, $2.92\pm0.31$ for seaweed, and $3.20\pm0.29$ for surf clam, compared with $2.26\pm0.34$ for salmon and $2.27\pm0.29$ for black rice. Its relative advantage thus persists on the held-out ingredients, although their lower absolute ratings indicate room to improve visual quality. AI-ensemble means tend to be higher than human means, but the direction and magnitude of the difference vary by method and category. 

\begin{table}[tbp]
\centering
\caption{Offline 33-frame future-vision reconstruction diagnostic on a fixed 176-clip set (5,808 frames/method). Lower is better. The visual readout is absent from deployed control, and these errors do not by themselves isolate latent quality.}
\label{tab:wam-eval}
\begin{tabular}{lccc}
\toprule
Model & MAE (\%) $\downarrow$ & $1-\mathrm{SSIM}$ (\%) $\downarrow$ & LPIPS $\downarrow$ \\
\midrule
M3 & 11.4\% & 35.1\% & 0.29 \\
M4 & 7.5\% & 28.9\% & 0.21 \\
M5 & 5.6\% & 22.9\% & 0.17 \\
\textbf{M6 (\method{})} & \textbf{3.8\%} & \textbf{17.7\%} & \textbf{0.13} \\
\bottomrule
\end{tabular}
\end{table}

\subsection{Manipulation Success and Controlled Comparisons}
Full \method{} achieves success rates of 75\%, 60\%, and 70\% on rice ball, seaweed, and surf clam, respectively (68.3\% ID average), and 35\% and 40\% on held-out black rice and salmon (37.5\% OOD average). Both factors improve average ID and OOD success when the other is held fixed. Enabling consequence supervision raises ID/OOD success from 15.0/5.0\% to 25.0/17.5\% with direct tactile concatenation (M3$\rightarrow$M4), and from 36.7/10.0\% to 68.3/37.5\% with gated fusion (M5$\rightarrow$M6). Replacing direct tactile concatenation with feature-wise gated fusion raises ID/OOD success from 15.0/5.0\% to 36.7/10.0\% without consequence supervision (M3$\rightarrow$M5), and from 25.0/17.5\% to 68.3/37.5\% with consequence supervision (M4$\rightarrow$M6).

These matched comparisons provide the main experimental evidence for the two design choices. M5 and M6 have identical deployed inputs and architecture, yet M6 achieves 31.7 percentage points higher ID success and 27.5 points higher OOD success. This comparison supports the benefit of training-only consequence supervision without additional deployment-time sensing or decoder computation. With consequence supervision enabled, M6 exceeds the concatenation variant M4 by 43.3 ID points and 20.0 OOD points, supporting the benefit of feature-wise gated tactile fusion over direct tactile concatenation. Since all four variants receive current touch, these comparisons assess how tactile information is fused, not whether it is available. The positive contrasts support complementary benefits of gated fusion and consequence supervision.

Full \method{} also exceeds the strongest system-level baseline, GR00T+tactile (M2), by 20.0 ID points, 22.5 OOD points, and 21 points in the descriptive five-task pool (56\% versus 35\%). Pooled success for M1--M6 is 15\%, 35\%, 11\%, 22\%, 26\%, and 56\%, respectively. The gains on black rice and salmon show transfer to the two prespecified ingredient shifts, but do not establish broad OOD generalization. Because M5 disables all consequence objectives together, its comparison with M6 attributes the benefit to their joint supervision rather than any individual prediction target.

\subsection{Auxiliary Future-Vision Diagnostic}
We next examine whether consequence supervision improves future-vision reconstruction. We evaluate the inherited visual readout offline for M3--M6 on a fixed diagnostic set of 176 synchronized clips, each containing $H_I=33$ target frames (5,808 frames per method). The diagnostic clips are trajectory-disjoint from all training episodes. M4 and M6 adapt the readout through future-vision supervision; M3 and M5 retain it for this diagnostic but receive no future-vision loss. Decoded RGB values are normalized to $[0,1]$. Table~\ref{tab:wam-eval} reports $100\times\mathrm{MAE}$, frame-averaged $100\times(1-\mathrm{SSIM})$ with an $11{\times}11$ Gaussian window~\cite{wang2004ssim}, and LPIPS~\cite{zhang2018lpips}; lower is better. Figure~\ref{fig:wam-prediction} shows qualitative examples.

With the deployed gated-fusion architecture held fixed, consequence supervision reduces the three errors from 5.6\%, 22.9\%, and 0.17 for M5 to 3.8\%, 17.7\%, and 0.13 for M6. The direct-concatenation pair M3/M4 shows the same trend. These results indicate improved future-vision reconstruction alongside the control improvements reported above. The predictions in Fig.~\ref{fig:wam-prediction} preserve coarse scene evolution but remain imperfect around rice and seaweed. This diagnostic does not isolate changes in $h_f$, because the future-vision objective updates both the shared representation and the decoder.

\section{Discussion}
The controlled comparisons assess two design choices: feature-wise gated fusion versus direct tactile concatenation, and the use of joint consequence supervision. Current tactile input is available in every Cosmos3 variant, so these ablations do not measure the benefit of adding touch to a vision-only policy. They also do not separate the effects of future vision, tactile summaries, task progress, contact risk, or failed-window supervision. Target-specific and stop-gradient ablations, evaluated with a common probe on trajectory-disjoint data, would distinguish these contributions.

Continual adaptation to new ingredients is a promising extension of this work. The benefit of training-only consequence supervision suggests that newly logged visual and tactile outcomes could help update contact-sensitive features without adding computation to the deployed action path. In particular, unsuccessful interactions could supply consequence targets while their commands remain excluded from imitation. A chronological adaptation study could test whether this strategy improves performance on new ingredients through fresh post-update rollouts, while replay and retention tests on the foods could assess preservation of learned skills. The present evaluation uses a fixed policy: none of the 600 evaluation rollouts is used for updates.

\vspace{-2pt}

\section{Conclusion}
We presented \method{}, a Cosmos3-based dexterous policy that uses feature-wise gated tactile fusion for control and action-conditioned future-consequence supervision during training. Failed windows retain their observed consequence targets while their actions are excluded from imitation, and all auxiliary readouts are removed at deployment. Across 600 real-robot rollouts, full \method{} achieved higher visual-outcome success than matched variants using direct tactile concatenation or omitting consequence supervision, including on two ingredients held out from task-specific training. Both human and AI evaluators assigned it the highest mean visual-quality score in every category, although scores were lower on held-out ingredients. These findings support combining gated tactile fusion with predictive supervision, while leaving their underlying mechanisms and consistency across training seeds for further study.

\vspace{-2pt}

\bibliographystyle{IEEEtran}
\bibliography{references}

\end{document}